\documentclass[10pt,twocolumn,letterpaper]{article}

\usepackage[pagenumbers]{wacv} 

\usepackage{graphicx} 
\usepackage{amsmath} 
\usepackage{amssymb} 
\usepackage{booktabs}
\usepackage{xspace}
\usepackage{times} 
\usepackage{epsfig} 
\usepackage{tabularx}
\usepackage{multirow}
\usepackage{multicol}
\usepackage{makecell}
\usepackage[shortlabels]{enumitem}
\usepackage{tabularx}
\usepackage[normalem]{ulem}
\usepackage{float}
\usepackage{graphicx}
\usepackage{enumitem}
\usepackage{wrapfig}

\usepackage[pagebackref,breaklinks,colorlinks]{hyperref}

\usepackage[capitalize]{cleveref}
\crefname{section}{Sec.}{Secs.}
\Crefname{section}{Section}{Sections}
\Crefname{table}{Table}{Tables}
\crefname{table}{Tab.}{Tabs.}

\def\wacvPaperID{2157} 
\def\confName{WACV}
\def\confYear{2024}

\begin{document}
\title{Segment anything, from space?}

\author{Simiao Ren\textsuperscript{1}
\and
Francesco Luzi*\textsuperscript{1}
\and
Saad Lahrichi*\textsuperscript{2}
\and
Kaleb Kassaw*\textsuperscript{1}
\and
Leslie M. Collins\textsuperscript{1}
\and
Kyle Bradbury\textsuperscript{1,3}
\and
Jordan M. Malof\textsuperscript{4}
\and
\textsuperscript{1} Electrical and Computer Engineering, Duke University\\
\textsuperscript{2} Division of Natural and Applied Sciences,
Duke Kunshan University\\
\textsuperscript{3} Nicholas Institute for Energy, Environment \& Sustainability, Duke University\\
\textsuperscript{4} Computer Science, University of Montana\\
{\tt\small \{simiao.ren, francesco.luzi, saad.lahrichi, kaleb.kassaw\}@duke.edu},\\
{\tt\small\{leslie.collins, kyle.bradbury\}@duke.edu, jordan.malof@umontana.edu}}





\def\eg{\emph{e.g}\bmvaOneDot}
\def\Eg{\emph{E.g}\bmvaOneDot}
\def\etal{\emph{et al}\bmvaOneDot}


\maketitle
\begin{abstract}
Recently, the first foundation model developed specifically for image segmentation tasks was developed, termed the "Segment Anything Model" (SAM).  SAM can segment objects in input imagery based on cheap input prompts, such as one (or more) points, a bounding box, or a mask.  The authors examined the \textit{zero-shot} image segmentation accuracy of SAM on a large number of vision benchmark tasks and found that SAM usually achieved recognition accuracy similar to, or sometimes exceeding, vision models that had been trained on the target tasks. The impressive generalization of SAM for segmentation has major implications for vision researchers working on natural imagery.  In this work, we examine whether SAM's performance extends to overhead imagery problems and help guide the community's response to its development. We examine SAM's performance on a set of diverse and widely studied benchmark tasks.  We find that SAM does often generalize well to overhead imagery, although it fails in some cases due to the unique characteristics of overhead imagery and its common target objects.  We report on these unique systematic failure cases for remote sensing imagery that may comprise useful future research for the community.  
\end{abstract}

\section{Introduction}

Foundation models are large deep learning models (e.g., in terms of free parameters) that have been trained on massive datasets, giving them the ability to generalize well to novel down-stream tasks (e.g., novel datasets, or prediction targets) with little or no additional training (i.e., so-called few-shot or zero-shot generalization).  Recent foundation models have been focused on natural language processing (NLP) tasks (e.g., BERT \cite{devlin2018bert}, GPT-3 \cite{brown2020language}), while some work also focused on text and imagery (e.g., CLIP \cite{radford2021learning} and ALIGN \cite{jia2021scaling}).  Recently the first foundation model primarily designed for Segmentation tasks was developed, termed the "Segment Anything Model" (SAM) \cite{kirillov2023segment}.  SAM is designed to output a segmentation mask for a given input image based upon one (or several) of the following input prompts: one (or more) points, a bounding box, or a segmentation mask (e.g., one that is coarse).  

\begin{figure}
\begin{center}
    \includegraphics[width=\linewidth]{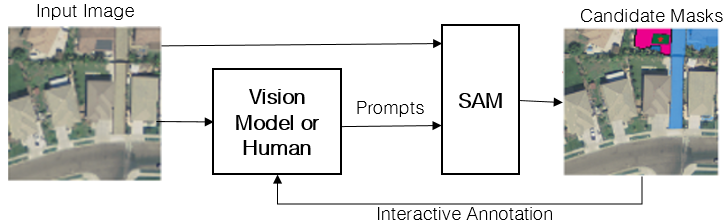}
\end{center}
\caption{Illustration of using SAM for segmenting solar arrays in satellite imagery.} 
\label{fig:sam_overview_diagram}
\end{figure}

The goal of SAM is to segment \textit{any} object in \textit{any} image only based upon these cheap input prompts, and without the need for additional task-specific or dataset-specific adaptation (e.g., training) \cite{kirillov2023segment}.  SAM's flexible input prompting \cite{kirillov2023segment}) and zero-shot segmentation scheme make it a highly flexible model that could be employed as a key component in a variety of vision systems.  To illustrate SAM's flexibility and potential impact, the authors in \cite{kirillov2023segment} demonstrated its effectiveness on several potential application scenarios.  Two important scenarios - which we aim to replicate in overhead imagery in this work - were \textit{interactive annotation} and so-called \textit{model composition}, which are broadly illustrated in Fig. \ref{fig:sam_overview_diagram} (see Sec. \ref{sec:composition_experiments} and Sec. \ref{sec:interactive_experiments} for more detail).

\textbf{Interactive Annotation.} In this scenario SAM is employed to reduce the cost of manual (human) annotation of target object instances.  Rather than drawing a mask by hand, the annotator provides cheap input prompts such as bounding boxes or point prompts to SAM.  The annotator can optionally provide additional prompts as feedback to SAM, to iteratively refine its segmentation, or \textit{mask}, predictions.  SAM was shown to generally outperform existing interactive annotation methods - often by a large margin - on natural imagery tasks and to be capable of producing highly accurate annotations with relatively little guidance \cite{kirillov2023segment}. 

\textbf{Model Composition.} In this scenario SAM is employed to predict a segmentation mask based on prompts from another vision model: e.g., a bounding box or another (potentially low-quality) mask. For example, given limited data, a detection model could be trained to produce a (potentially noisy) bounding box, which could then be used to prompt SAM.  In this scenario, and on natural imagery, SAM often achieved segmentation accuracy that was comparable to existing models that had been trained specifically for the target task (i.e., with full target label supervision) \cite{kirillov2023segment}.


\subsection{Contributions of this Work}
\begin{figure*}
    \centering
    \begin{tabular}{c}
    \quad \quad \includegraphics[width=0.85\textwidth]{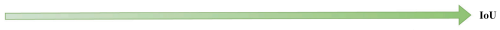}
    \end{tabular}
    \begin{tabular}{ccccccc}
         &  & IoU: 0.165  &  & IoU: 0.475  & & IoU: 0.952\\
        \parbox[t]{4mm}{\rotatebox[origin=l]{90}{\quad Building}} & \includegraphics[width=0.14\textwidth]{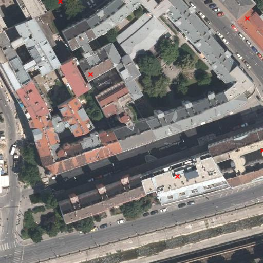} &  \includegraphics[width=0.14\textwidth]{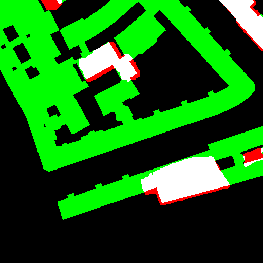} & \includegraphics[width=0.14\textwidth]{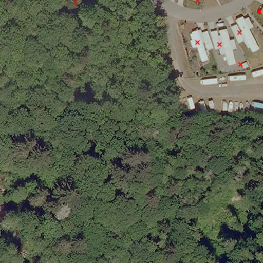} & \includegraphics[width=0.14\textwidth]{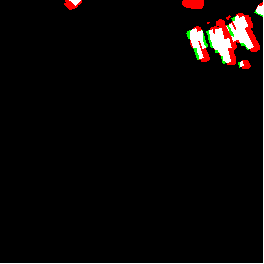} &
        \includegraphics[width=0.14\textwidth]{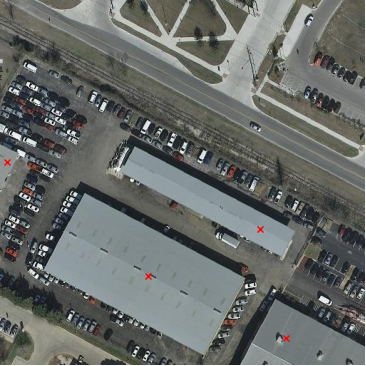} &
        \includegraphics[width=0.14\textwidth]{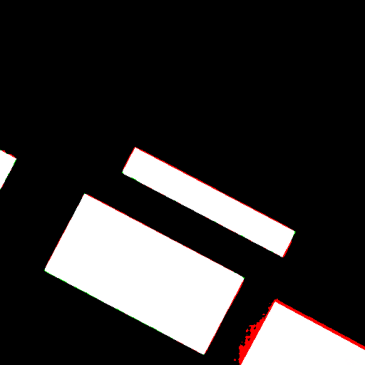}\\
         &  & IoU: 0.003  &  & IoU: 0.503  & & IoU: 0.754\\
        \parbox[t]{4mm}{\rotatebox[origin=l]{90}{\quad \quad Road}} & \includegraphics[width=0.14\textwidth]{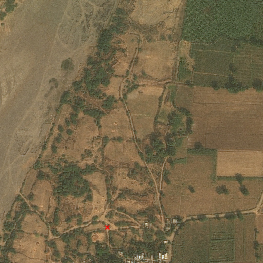} &  \includegraphics[width=0.14\textwidth]{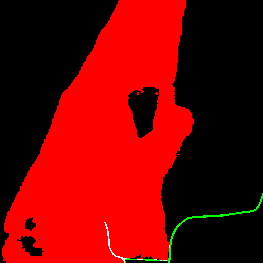} & \includegraphics[width=0.14\textwidth]{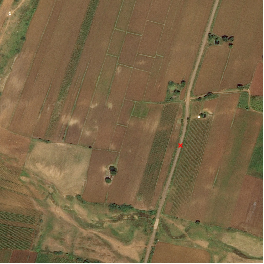} & \includegraphics[width=0.14\textwidth]{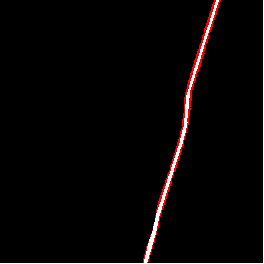} &
        \includegraphics[width=0.14\textwidth]{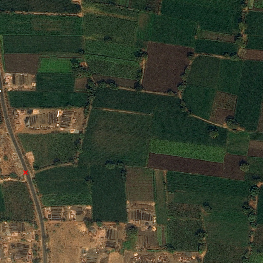} &
        \includegraphics[width=0.14\textwidth]{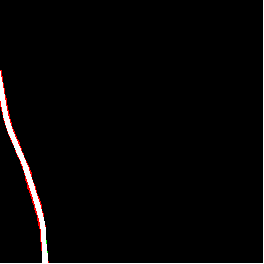}\\
         &  & IoU: 0.104  &  & IoU: 0.429  & & IoU: 0.956\\
        \parbox[t]{4mm}{\rotatebox[origin=l]{90}{\quad \quad Solar}} & \includegraphics[width=0.14\textwidth]{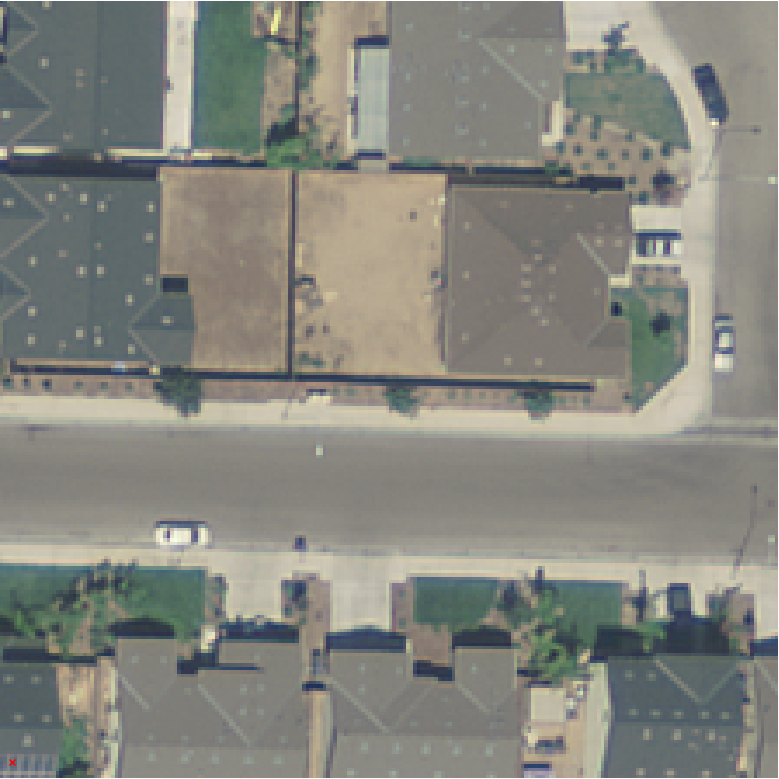} &  \includegraphics[width=0.14\textwidth]{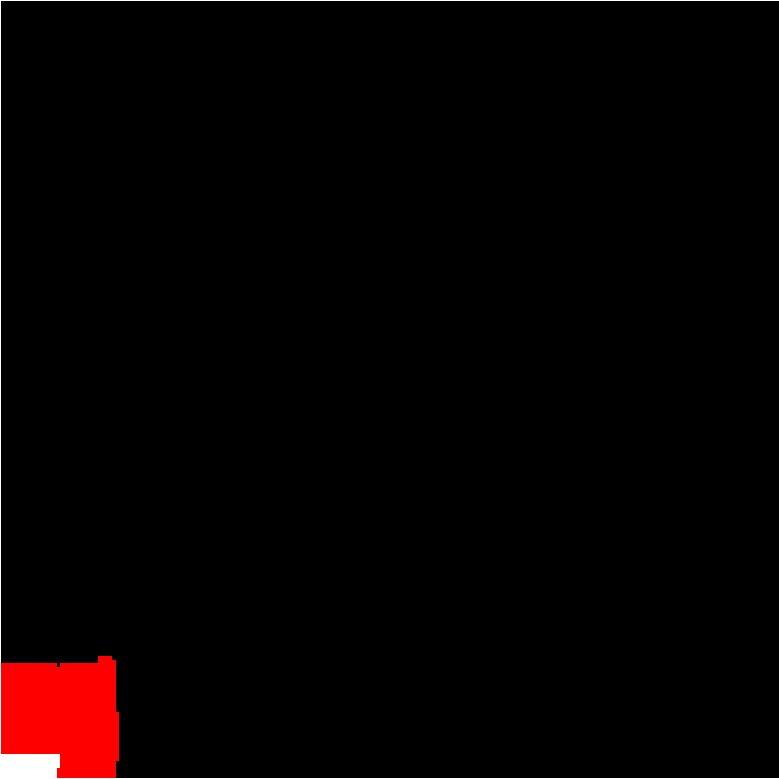} & \includegraphics[width=0.14\textwidth]{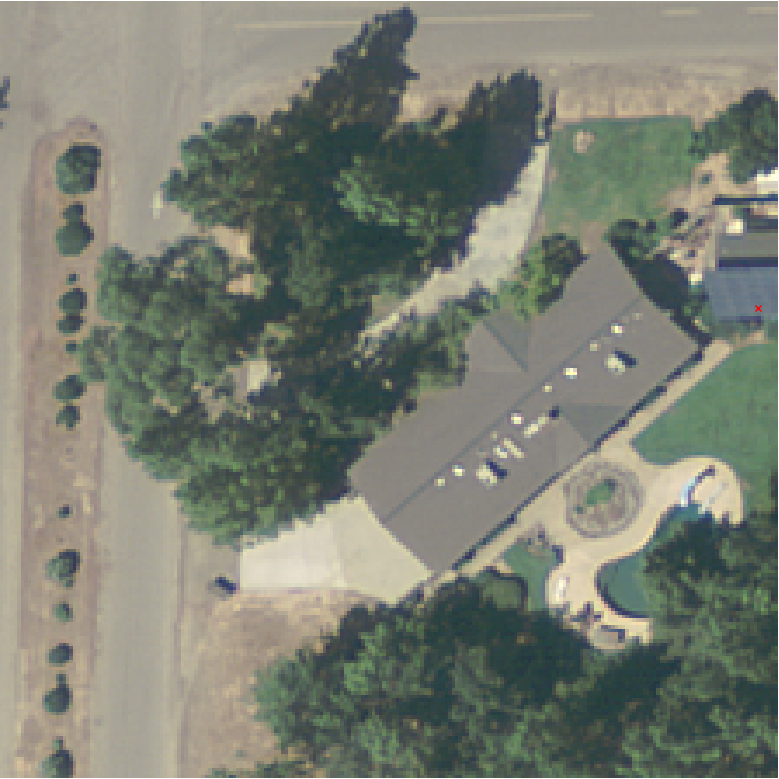} & \includegraphics[width=0.14\textwidth]{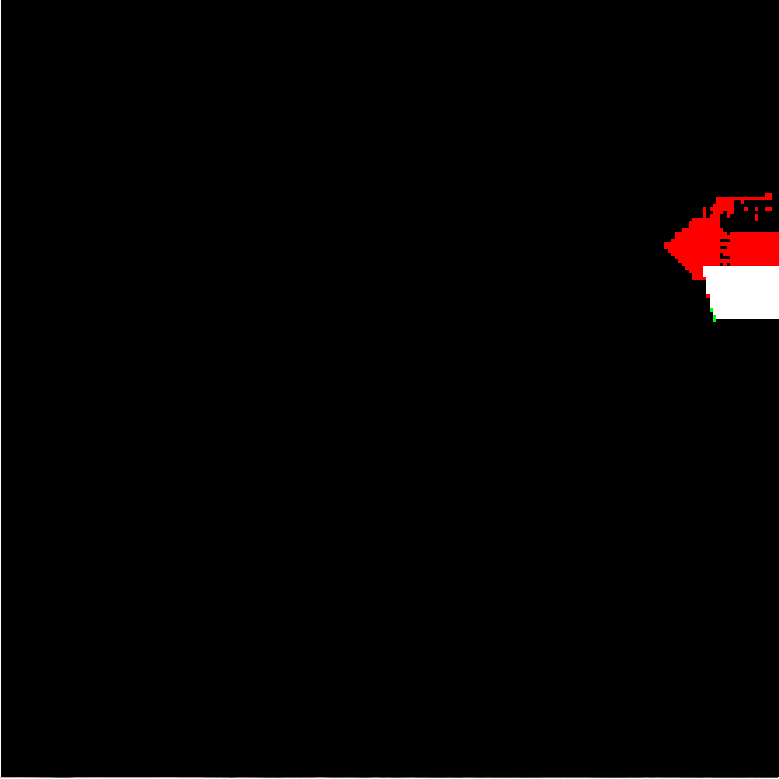} &
        \includegraphics[width=0.14\textwidth]{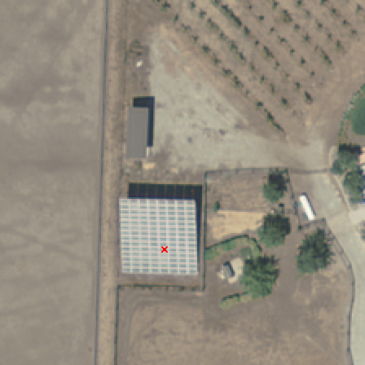} &
        \includegraphics[width=0.14\textwidth]{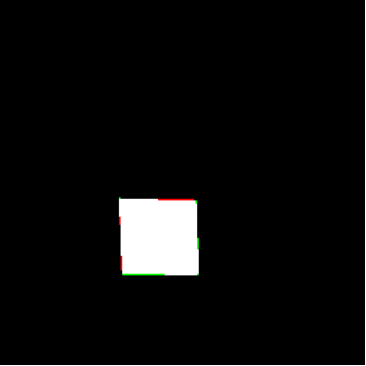}\\
         &  & IoU: 0.040  &  & IoU: 0.509  & & IoU: 0.841\\
        \parbox[t]{4mm}{\rotatebox[origin=l]{90}{\quad \quad Crop}} & \includegraphics[width=0.14\textwidth]{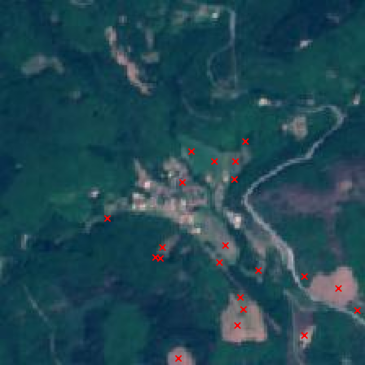} &  \includegraphics[width=0.14\textwidth]{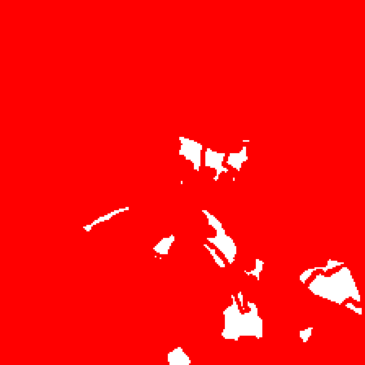} & \includegraphics[width=0.14\textwidth]{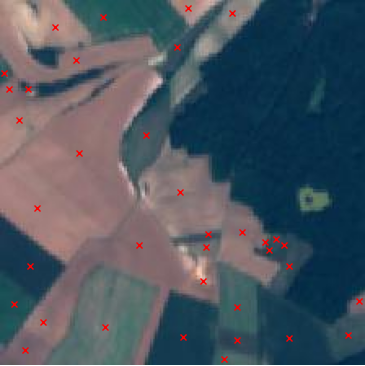} & \includegraphics[width=0.14\textwidth]{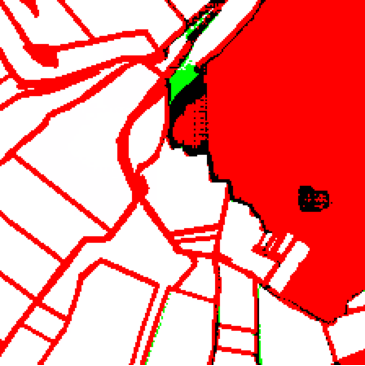} &
        \includegraphics[width=0.14\textwidth]{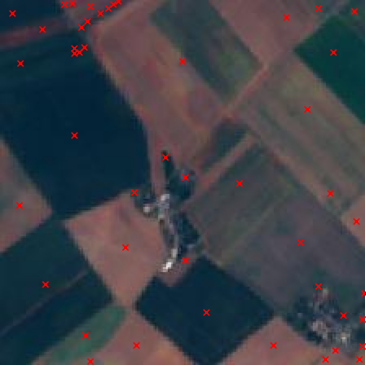} &
        \includegraphics[width=0.14\textwidth]{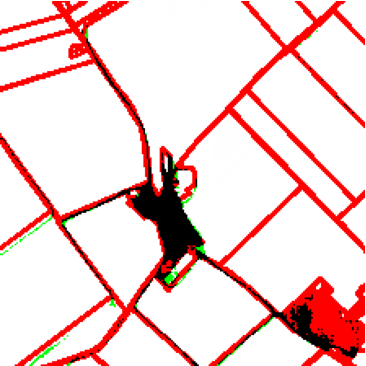}\\
    \end{tabular}
    
    \caption{The input image, prompts, and segmentation output for SAM is shown for the building, road, solar panel, and crop segmentation tasks. Each row displays a specific segmentation task and the input output image pairs are ordered from worst performing to best performing. Red X's in the input images represent the prompts given to SAM. The output is shown in three colors where white represents the correctly predicted pixels, red shows the false positive pixels, and green shows the missed pixels. The image-wise IoU is shown above each output image.}
    \label{fig:test}
\end{figure*}

The authors of SAM reported comprehensive empirical evidence demonstrating the effectiveness of SAM for natural imagery, however, they only investigated a single overhead imagery dataset.  Vision problems for overhead imagery exhibit unique challenges compared to natural imagery, and therefore it is unclear whether, or to what extent, the effectiveness of SAM transfers to tasks involving overhead imagery.  In other words, we ask whether SAM can segment anything \textit{from space}.  For example, Fig. \ref{fig:test} presents some out-of-the-box results with SAM on varying tasks, illustrating a variety of performance levels.  To answer this question, we replicate the experiments from \cite{kirillov2023segment} that correspond to two important potential applications in overhead imagery: interactive annotation and model composition.  

We adapt the aforementioned experiments of \cite{kirillov2023segment} to seven existing public benchmark datasets of overhead imagery  (see Table \ref{tbl:dataset_description}), encompassing 5 million square kilometers of surface area. The benchmarks include a variety of widely-studied object classes (e.g., buildings, roads, cloud cover, farming crops), image resolutions (e.g., 0.3m - 30m), and geographic locations (e.g., including Africa, Europe, Asia, and North America).  To our knowledge, this is the first systematic and comprehensive investigation of SAM on overhead imagery, including experiments relevant to both model composition and interactive annotation, thereby providing valuable guidance to the research community. From our experiments, we also identify SAM failure cases that are unique to overhead imagery and suggest potential future areas of research for the community. 

\section{Related Work}
In the short time since its publication, SAM has been used in numerous applications including image dehazing \cite{jin2023let}, image tagging \cite{zhang2023recognize}, shadow segmentation \cite{wang2023detect}, sonar segmentation \cite{wang2023sam}, foreground segmentation \cite{xu2023aquasam}, electron microscopy segmentation \cite{cheng2023axoncallosumem, larsen2023np}, and many medical segmentation applications \cite{hu2023efficiently, wang2023sam, yue2023surgicalsam, kim2023empirical, li2023single, horst2023cellvit, ahmadi2023comparative, he2023accuracy}. There have also been several applications of SAM in remote sensing, including segmentation of geological features and landforms in planetary structures \cite{julka2023knowledge}, segmentation of glaciers \cite{stearns2023segment} and sea ice \cite{yu2023sea}, road segmentation \cite{ji2023segment}, and building segmentation \cite{ji2023segment, zhang2023text2seg}.  Our work represents the first comprehensive evaluation of SAM on overhead imagery (e.g., including a variety of target classes, resolutions, and geographic locations), and the first to include both model composition and interactive annotation scenarios. Collectively, therefore, our results are the first that assess how well SAM transfers to overhead imagery applications.

\section{Benchmark Datasets}
\label{sec:dataset_description}
For our experiments, we utilized eight benchmark datasets of overhead imagery that were selected for inclusion based on several criteria.  We first had two strict criteria for inclusion: (i) the availability of class-level or instance-level segmentation labels; and (ii) the datasets were publicly available, to enable further study by the community.  Among the datasets that satisfied these criteria, we also applied several softer criteria for inclusion: (i) we prioritized datasets that were of greater interest to the overhead imagery community, as evidenced by their inclusion in many prior publications; (ii) benchmarks that involved widely-studied target objects (e.g., buildings, roads, land use); and (iii) datasets that collectively resulted in a representative set of key properties (e.g., geographic location, image resolution, or target classes).  The resulting set of eight datasets that we selected is shown in Table \ref{tbl:dataset_description}, along with references and key details.  Further details about our benchmark datasets can be found in the Supplement. 

\textbf{Additional Data Preparation Details.}The datasets we ultimately used for experimentation were constructed based upon the benchmarks in Table \ref{tbl:dataset_description}.  In many cases, we used the datasets without modification: Solar, 38-Cloud, DeepGlobe Roads, SpaceNet2.  However, we made some modifications to other datasets. Following a recent large-scale comparison of building segmentation models \cite{luzi2023transformers}, we combine the Inria and the DeepGlobe Building datasets into a single dataset, \textit{DG+Inria Building}, allowing us to compare our results with SAM to the strong supervised models from \cite{luzi2023transformers}. The DeepGlobe Land dataset comprises seven classes, and all pixels are assigned to one of seven classes, resulting in a large number of potential object instances; in our experiments, we treat each connected component within each class as an instance (see Sec. \ref{sec:composition_experiments}). To make the problem less computationally intensive, we chose a subset of three classes of increasing visual complexity (water, agriculture, and urban) and in each case, we set their labels to one, and all other classes to zero, resulting in three separate problems: \textit{DG-land-water}, \textit{DG-land-urban}, \textit{DG-land-agri}. Lastly, the Parcel Delineation dataset is natively an edge detection dataset; however, we designated each isolated component of non-edge pixels as an object instance.  


\setlength{\tabcolsep}{2pt}
\begin{table}[tp]
  \begin{center} 
           \scalebox{.85}{
        \begin{tabular}{|c|c|c|c|c|c|} 
            \toprule 
            Dataset  & Unique & Classes  & Size  & Resolution  & Task \\ 
              & Locations &   &  (km$^2$) &  (m) & Type  \\ 
            \midrule 
            \midrule
            Solar \cite{bradbury2016distributed}    & 4 &   Solar Panel  &  1,352.25 & 0.30 & Seg \\ 
            \hline
            Inria \cite{maggiori2017dataset}     &   5&  Building &  405  &  0.30 & Seg\\ 
            \hline
            DeepGlobe \cite{demir2018deepglobe}  &   4&   Building   &  398.28  &  0.31  & Seg\\ 
            \hline
             38-Cloud \cite{38-cloud-2} &  NA    & Cloud   &  2,188,800   & 30 & Seg\\ 
            \hline
            DeepGlobe    & \multirow{2}{*}{3}  &    \multirow{2}{*}{Roads}  &  \multirow{2}{*}{2,220}  &  \multirow{2}{*}{0.50} & \multirow{2}{*}{Seg} \\ 
            Roads \cite{demir2018deepglobe}   &  & & & & \\ 
            \hline
             Parcel       &  \multirow{2}{*}{1}   &  Crop   &  \multirow{2}{*}{4,403.84}  & \multirow{2}{*}{10}  & \multirow{2}{*}{Edge} \\ 
             Delineation  \cite{aung_farm_2020}     &  & Boundaries & & & \\
             \hline
             DeepGlobe & \multirow{2}{*}{3} & Land &  \multirow{2}{*}{1,716.9} & \multirow{2}{*}{0.50} & \multirow{2}{*}{Seg} \\
             Land \cite{demir_2018_deepglobe} &  &  Use &  &  & \\
            \hline 
             SpaceNet 2 \cite{van2018spacenet} & 4 &  Building & 3011 & 0.30 & Seg \\
            \bottomrule 
        \end{tabular}
    }
    \end{center} 
\vspace{-0.5em}
\caption{A summary of the quantity, resolution, and type of data we evaluate. The resolution is in units of meters per pixel. For task type, we have "Seg" for segmentation and "Edge" for edge detection.} 
\label{tbl:dataset_description}
\end{table} 
\setlength{\tabcolsep}{6pt}

\section{Model Composition Experiments}
\label{sec:composition_experiments}
\begin{figure}[h!]
\centering
\includegraphics[width=0.9\linewidth]{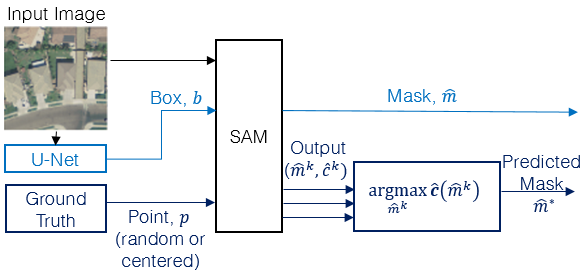}
\caption{Illustration of the Model Composition experimental design, adapted from \cite{kirillov2023segment} to our Overhead Imagery application. Please see text for detailed description.}
\label{fig:composition_experiment_diagram}
\end{figure}

In the Composition scenario, SAM is utilized to enhance the predictions made by other vision models.  Specifically, following \cite{kirillov2023segment}, we assume that SAM is prompted with either a single point, $p$, or a bounding box, $b$, that has been produced by some other vision model.  The goal of SAM is to produce an accurate instance mask based upon these simpler, and potentially imperfect, prompts.  To emulate the desired model-based prompts, we train U-Net models to generate the bounding box prompts, and we use the ground truth labels from each benchmark to generate the point-based prompts.  Our experimental design is illustrated in Fig \ref{fig:composition_experiment_diagram}.  

\textbf{U-Net Models.} We train a U-Net segmentation model for each of our benchmark datasets, denoted \textit{U-Net (ours)}.  Note that most of our benchmarks have class-level segmentation labels (as opposed to instance-level) and therefore we train our models to predict class-level masks.  For each benchmark we create disjoint training and testing data partitions for model training and evaluation.  Where possible we use pre-established partitions for each benchmark (e.g., Inria, DeepGlobe, SpaceNet).   For some of our benchmarks - where they are available - we also report the IoU of a recent State-of-the-Art (SOTA) U-net model, denoted \textit{U-Net (SOTA)}.  Full details of these models can be found in the Supplementary Materials (e.g., architecture, training procedures, source publications).  These models reflect the current performance that can be achieved using fully supervised techniques, providing a useful comparison to SAM.  

\textbf{Bounding Box Prompts.}  Based upon the output of the class-level masks output by the \textit{U-Net (ours)} model, we extract instance-level bounding boxes that can then be used to prompt SAM.  This is achieved by treating each connected component in the U-Net output as an instance-level mask.  Then for each of these instance-level masks, we compute the smallest bounding box that encloses it and use the resulting box to prompt SAM.  From these prompts, SAM produces instance-level mask predictions, denoted $\hat{m}_{i}$, where $i$ refers to the mask generated by the $i^{th}$ box prompt.  To evaluate the accuracy of SAM's masks, we use them to create a single class-level mask $\hat{m} = \cup_{i} \hat{m}_{i}$, which can be scored against the class-level ground truth masks available with our benchmark datasets.  

\textbf{Point Prompts.} Our point-based prompts are generated using the ground truth labels that are available with each benchmark dataset.  Except for SpaceNet, our benchmark datasets all provide class-level labels.  We treat connected components in the ground truth masks as instance-level masks.  Within each mask, we generate two point prompts: one by selecting a random point within the mask, and one by selecting the center point in the mask (where "center" refers to the point most distant from the mask boundary). We score SAM's performance once using only the random points, and once only the center points.  When SAM is prompted with a single point it produces three candidate masks, denoted $\hat{m}^{k}$, and a score for each mask indicating the model's estimate of the IoU of that mask with the true underlying object, denoted $\hat{c}^{k}$. SAM returns as a prediction the mask with the highest IoU prediction, which we denote $\hat{m}^{*}$.  Similar to bounding boxes, we produce a single class-level mask by taking the union of all instance-level mask predictions.

\setlength{\tabcolsep}{2pt}
\setlength{\tabcolsep}{6pt}

\begin{figure*}[h!]
\centering
\includegraphics[width=\linewidth]{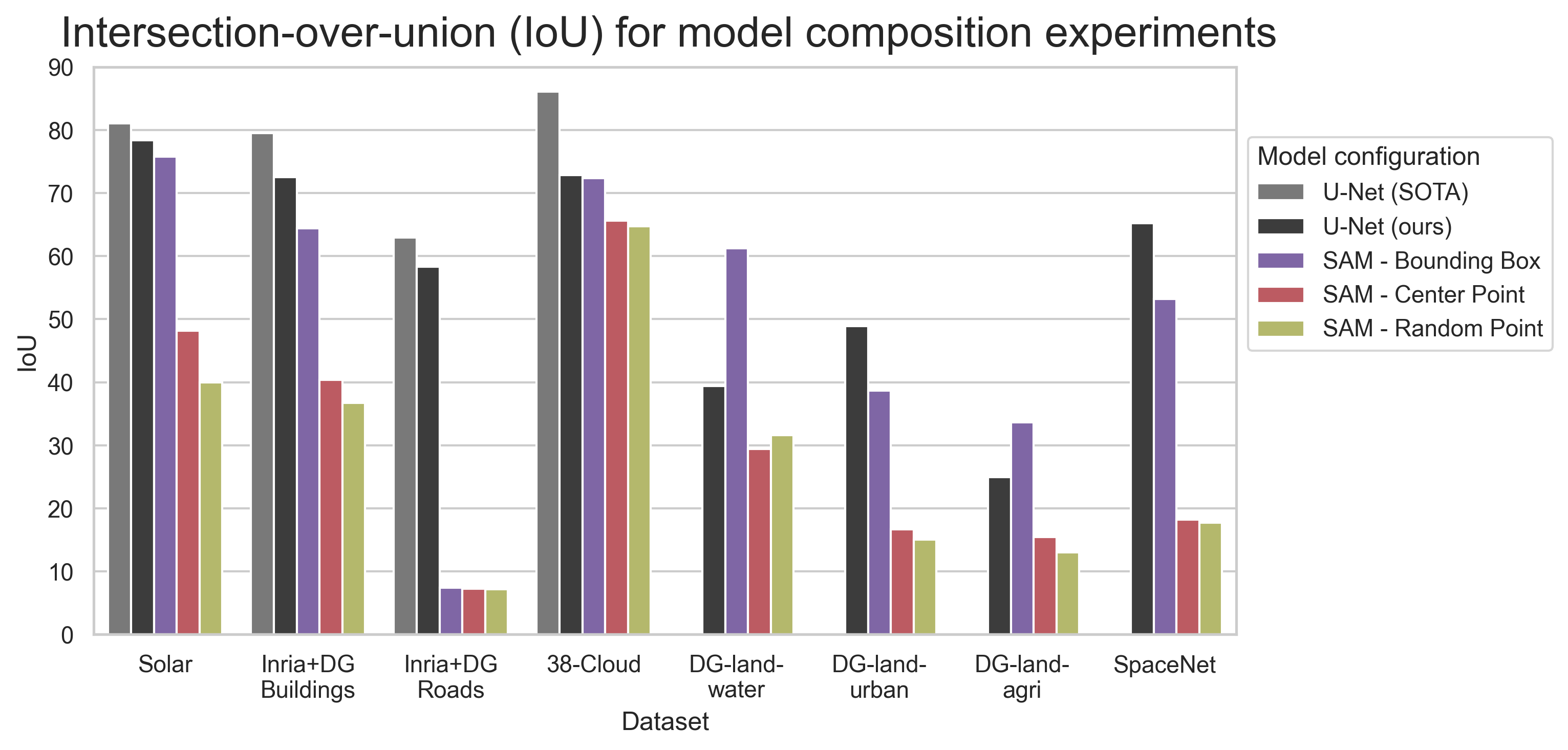}
\caption{The intersection-over-union for all models included in our Model Composition experiments.}
\label{fig:results_composition}
\end{figure*}

\subsection{Results and Discussion}
The results of our composition experiments are presented in Fig. \ref{fig:results_composition}. The results indicate that the task-specific supervised models almost always achieve the highest IoU.  This is unsurprising since these models were trained using a large quantity of task-specific segmentation masks.  The \textit{U-Net (SOTA)} model always outperforms the \textit{U-Net (ours)}, reflecting the additional effort invested in its specialized design for each benchmark. Therefore, the IoU of the supervised models represents the performance that can be achieved using contemporary vision models out-of-the-box - in the case of \textit{U-Net (ours)} - and contemporary vision models with significant design investment - in the case of \textit{U-Net (SOTA)} model.  We also see that SAM consistently performs significantly worse with point prompts compared to more informative (but more expensive) box prompts, and that the position of point prompts (center versus random) has little effect on results. These findings for overhead imagery are largely consistent with those reported for SAM for natural imagery in \cite{kirillov2023segment} (e.g., on COCO and LVIS), where SAM's performance was also always inferior to a supervised model, and less-informative prompts performed worse.  However, we highlight a few key differences in our findings here.  

One key difference in our findings is that the gap in performance between SAM and supervised models (the U-Net models in our case) is significantly larger on most of our benchmarks. This is especially noticeable on the largest and most widely studied benchmark target objects, such as buildings and roads. We hypothesize that this reflects SAM's bias towards objects in natural imagery.  For example, the building class is highly complex (e.g., buildings are often large, and comprise many object-like sub-components), and therefore they likely cannot be well-characterized using a generic "objectness" concept inferred based upon natural imagery. However, SAM also performs relatively poorly (compared to supervised models) on solar arrays, despite their apparent visual simplicity, suggesting that it is not guaranteed to generalize to overhead imagery even when the target objects are relatively simple.  As we discuss in \textbf{Sec. \ref{sec:additional_analysis}}, SAM's performance can depend strongly upon the scale of the objects (i.e., we find simply artificially up-sampling the imagery can be beneficial), potentially impacting its performance on solar arrays or other small object instances. Lastly, we see that SAM achieves substantially lower performance, and low performance overall, on the road class.  This highlights a major challenge with some target classes in overhead imagery where the notion of an object instance is not well-defined. We discuss this issue further in \textbf{Sec. \ref{sec:additional_analysis}}. 

Overall our results suggest that SAM can sometimes generalize well to overhead imagery tasks, providing competitive zero-shot results compared with supervised models in some cases.  However, in most cases, it performs relatively poorly compared to supervised models, especially heavily-engineered models (e.g., on buildings and clouds), and sometimes it completely fails (e.g., roads). Many of these challenges may be overcome by fine-tuning SAM on overhead imagery classes (e.g., buildings, clouds, land use), or by producing class-level segmentation rather than instance-level segmentations in cases where an instance concept is poorly defined (e.g., roads).

\section{Interactive Annotation Experiments}
\label{sec:interactive_experiments}

\begin{figure}[h!]
\centering
\includegraphics[width=0.9\linewidth]{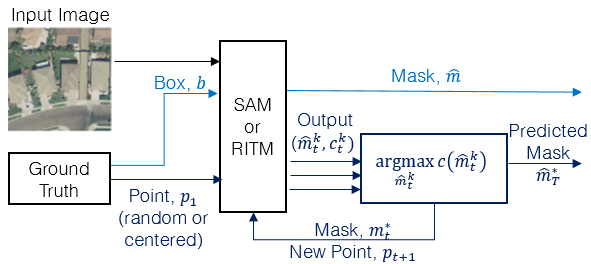}
\caption{Illustration of the experimental designs adopted from \cite{kirillov2023segment} to evaluate SAM for human-interactive annotation. We consider two general interactive annotation scenarios: a single bounding box prompt (light blue), and multi-point prompts (dark blue). In the first scenario (light blue) SAM is provided with a single bounding box prompt, denoted $b$ and it returns a single estimated mask, denoted $\hat{m}$.  In the second scenario (dark blue), SAM is initially provided with a single point prompt, denoted $p_{1}$, and returns as output three mask-score pairs, $(\hat{m}_{i}^{k}, \hat{c}_{i}^{k})$ each with an estimate of the IOU obtained.}
\label{fig:human_experiment_diagram}
\end{figure}

In the Interactive Annotation scenario, SAM is employed to reduce the time and effort required for a human to draw instance-level segmentation masks.  Rather than drawing a full mask by hand, the annotator can provide one (or more) cheap input prompts, from which SAM can provide richer instance-level masks.  Specifically, following \cite{kirillov2023segment}, we assume that SAM is prompted with either a series of points, denoted $p_{i}$ for the $i^{th}$ point; or a bounding box, denoted $b$.  In contrast to the composition scenario in Sec. \ref{sec:composition_experiments}, here we assume that the prompts are provided by a human annotator and this changes some of the experimental design.   Our experimental design, which follows \cite{kirillov2023segment}, is illustrated in Fig \ref{fig:human_experiment_diagram} and described below.  

\textbf{Bounding Box Prompts.}  Bounding box prompts are created in the same fashion as described in Sec. \ref{sec:composition_experiments} except that we use ground truth class-level masks available with our benchmark datasets instead of those generated by a supervised model.  Ground truth masks are generally more accurate than those generated by a model, reflecting the superior accuracy of a human annotator. Recapitulating this procedure, we extract instance-level prompts from the ground truth of each dataset, which are then used to prompt SAM.  From these prompts, SAM produces instance-level mask predictions, denoted $\hat{m}_{i}$, where $i$ refers to the mask generated by the $i^{th}$ box prompt.  To evaluate the accuracy of SAM's masks, we again use them to create a single class-level mask $\hat{m} = \cup_{i} \hat{m}_{i}$, which can be scored against the class-level ground truth masks available with our benchmark datasets.  

\textbf{Single Point Prompts.} In these experiments we evaluate SAM's performance when prompted with a single point, similar to Sec. \ref{sec:composition_experiments}.  For this purpose we use exactly the same procedure for generating the point prompts as was described in Sec. \ref{sec:composition_experiments}.  In summary, we extract instance-level masks from the ground truth masks available for each of our benchmark datasets and then prompt SAM either with (i) a random point, or (ii) a centered point, in each instance-level mask.  When prompted with a point, SAM produces three candidate masks and corresponding estimates of their IoU, $\hat{m}^{k}_{t},\hat{c}^{k}_{t})$, where $k$ indexes over the three masks, $t$ indexes the iteration of interactive segmentation ($t=1$ in the single point case). In the composition experiments in Sec. \ref{sec:composition_experiments} we selected the mask with the largest $c$ value as the final mask, however, a human annotator can evaluate the \textit{true} IoU of each candidate mask, denoted $c^{k}_{t}$. Therefore, in this case, we select as the output mask, the one that has the highest true IoU, denoted $\hat{m}^{*}$.  Once we have $\hat{m}^{*}$, we aggregate SAM's instance-level mask predictions in the same way as Sec. \ref{sec:composition_experiments} and report the resulting IoU for each benchmark.  Also following \cite{kirillov2023segment} we compare to the RITM interactive annotation model \cite{sofiiuk2022reviving}, which is evaluated using the same random and centered point prompts, respectively, that were used to prompt SAM. 

\textbf{Interactive Point Prompts.} The goal of this experiment is to emulate an interactive point-prompting scenario, in which a human annotator provides a series of point prompts, and where each point is used to improve upon the masks generated using all the previous prompts.  Specifically, and following \cite{kirillov2023segment}, we begin by prompting SAM with an initial point, $p_{1}$, which is obtained using the center point prompt from the Single Point Prompt above.   The final mask chosen for the next iteration of annotation is the one with the highest true IoU, denoted $\hat{m}^{*}_{t}$.  Based upon $\hat{m}^{*}_{t}$, another point prompt, $p_{t+1}$ is generated with the following process: we identify the largest contiguous error regions (either false positive or false negative) and then produce a point prompt in the center of that region.  The point prompt also assigned a number, which is input to SAM, indicating whether the point corresponds to a false negative or false positive region.  Then we input $p_{t+1}$ and $\hat{m}_{t}^{*}$ to SAM to generate another set of candidate masks.  We evaluate the performance of SAM and RITM as a function of $T$, the total number of iterations (and total number of point prompts) provided by SAM.  

\subsection{Results and Discussion}

\begin{figure*}[ht]
\centering
\includegraphics[width=\linewidth]{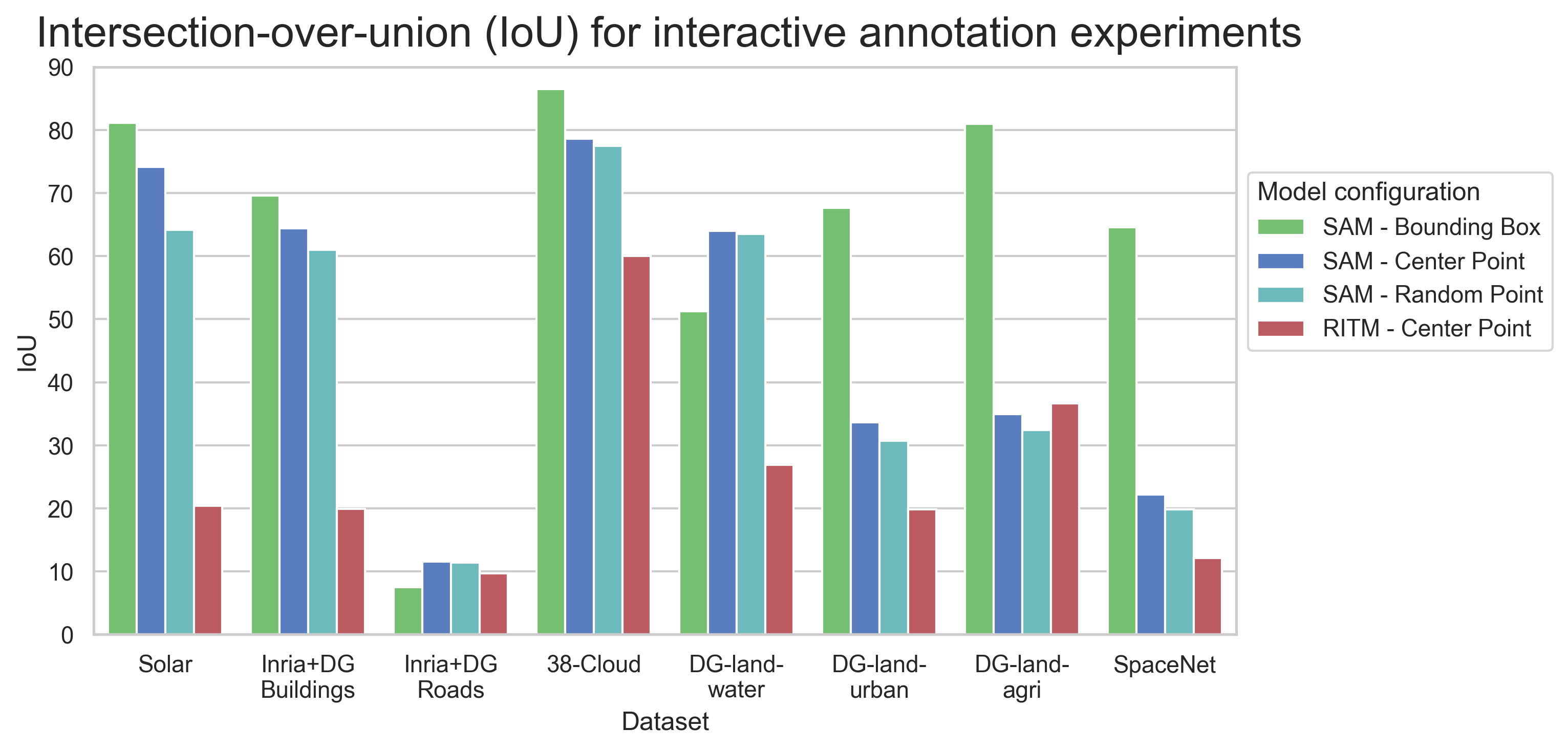}
\caption{Results of (pixel) intersection-over-union scores for each interactive annotation scenario.}
\label{fig:results_interactive}
\end{figure*}

\begin{figure}[ht]
\centering
\includegraphics[width=\linewidth]{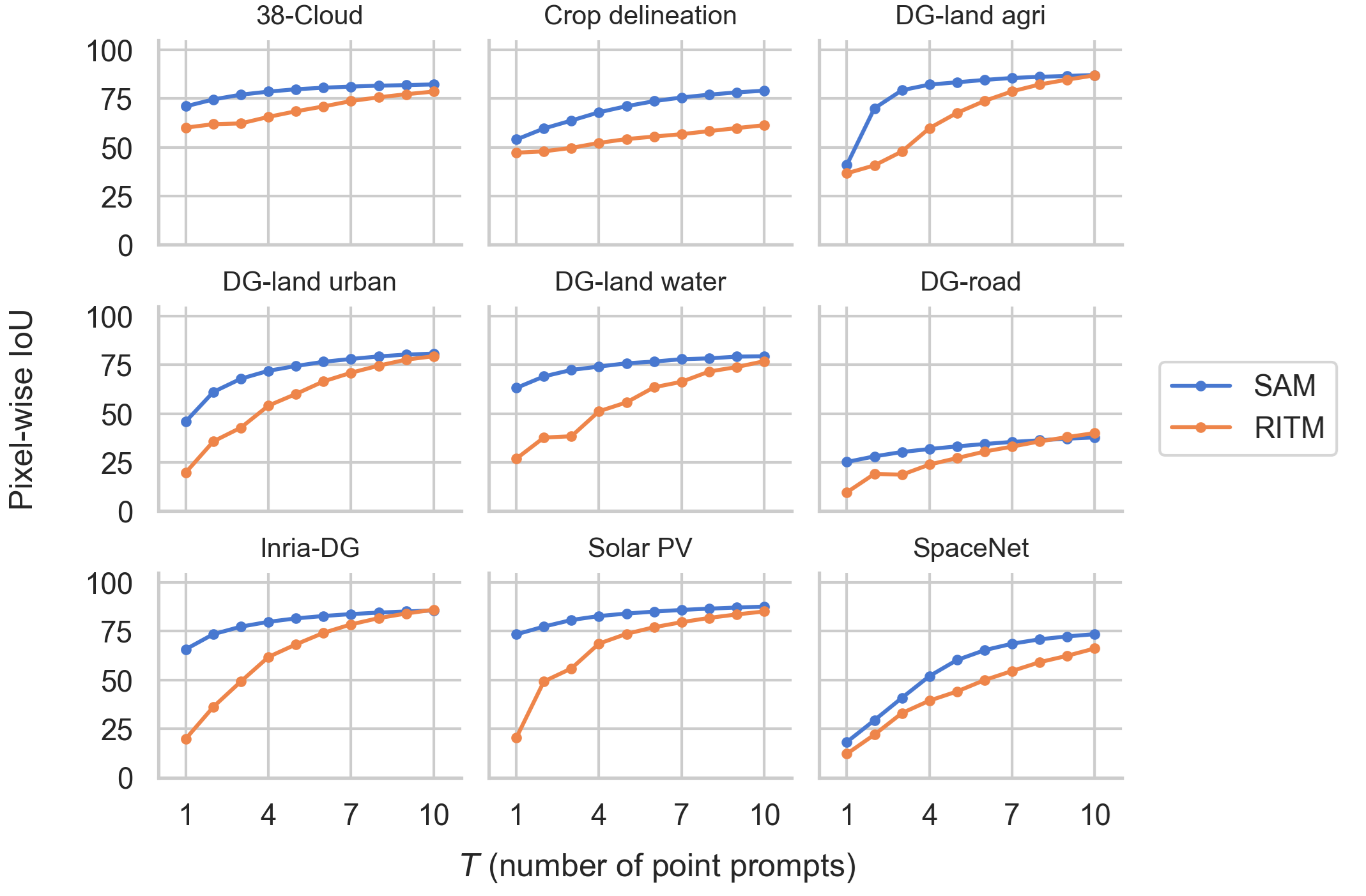}
\caption{Results of iterative prompts.}
\label{fig:results_iterative}
\end{figure}

The results of our single-point and single-box prompt experiments are reported in Fig. \ref{fig:results_interactive}.  The results indicate that the IoU of SAM tends to decrease as the informativeness of the prompts decreases, where bounding boxes are most informative, and random center points are least informative.  SAM seems relatively insensitive to whether the prompt points are centered or random.  We also find that SAM almost always outperforms RITM, and often by a substantial margin.  For example, on the Solar and Inria+DG Buildings datasets, SAM outperforms RITM by 50 and 45 IoU points, respectively.  These results are largely consistent with those reported for natural imagery in \cite{kirillov2023segment}, although the advantage of SAM over RITM seems even larger on overhead imagery than on natural imagery.  Also, notably, and similar to our model composition experiments in Sec. \ref{sec:composition_experiments}, all approaches perform very poorly on Inria+DG Roads, which we discuss in Sec \ref{sec:additional_analysis}.    

The results of our multi-point prompt experiments are reported in Fig. \ref{fig:results_iterative}.  The results indicate that the IOU of both SAM and RITM consistently improves as the number of prompt points increases. Notably SAM outperforms RITM in nearly every combination of benchmark or number of points, and SAM typically has the greatest performance advantages when there is a small number of points.  Once $T=10$ both approaches typically have similar IoU, and often achieve high overall IoUs that we expect would often be satisfying for extraction of ground truth masks (e.g., SAM achieves an IOU of 0.75, or larger, on 6 out of our 10 benchmarks when T=10). These results are largely consistent with those reported for SAM on natural imagery \cite{kirillov2023segment}, except that the overall IoUs are lower.  Collectively these results suggest that SAM offers state-of-the-art interactive annotation accuracy, and also often (though not always) achieves sufficiently high IoU to be a useful tool for human-interactive instance annotation.  Similar to our findings in the model annotation case, we hypothesize that substantial improvements in IoU could be obtained through fine-tuning SAM for overhead imagery, or for class-level segmentation.


\section{Additional Analysis}
\label{sec:additional_analysis}


In this section we discuss additional findings based on the collective experimental results in Sec. \ref{sec:composition_experiments} and Sec. \ref{sec:interactive_experiments}.

\begin{figure}[ht]
\centering
\includegraphics[width=\linewidth]{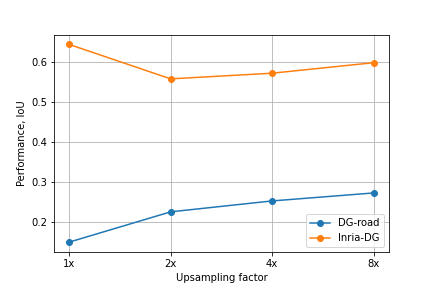}
\caption{Results of iterative prompts.}
\label{fig:results_scale}
\end{figure}

\textbf{Sensitivity of SAM to Image Resolution.} One finding of SAM was that it suffers from some sensitivity to image resolution.  It is well-known that using higher-resolution overhead imagery will enable higher accuracy for image recognition models, which is often attributed to the greater information content in the imagery.  However, we found that simply up-sampling our overhead imagery (i.e., adding no new information) can significantly improve SAM's IoU.  Fig. \ref{fig:results_scale} shows the IoU of SAM as a function of up-sampling factor for two benchmark problems, where in each case we artificially up-sampled the imagery.  The results indicate that the impact of up-sampling varies, but that SAM is generally sensitive to the number of pixels per unit of ground area, even if the information content in the image has not changed. We hypothesize that SAM may have a bias in its expected size of object instance.  In particular, SAM was trained on very high-resolution natural imagery, which in many cases biases it towards more pixels per object.  

\textbf{Instance Segmentation is Sometimes Ill-Posed.}  SAM is designed to perform object instance segmentation, however, some classes in overhead imagery may not be well-conceptualized in this manner. Fundamentally this problem can arise if there is no clear definition of an instance of a target class, or the definition may vary significantly across geography or application areas.  One example is roadways, a widely-studied problem in overhead imagery \cite{buslaev2018fully, zhou2018d, zhu2021global, chen2022road}, which is challenging because roads are large objects that are spatially connected over very large geographic regions.  Therefore it is unclear how to define when one road instance ends, and another begins.  Consequently, SAM consistently performs very poorly (IoU$<$10) on the road segmentation task (i.e., Inria+DG Roads) in all of our experiments compared to models that have been trained for class-level segmentation, which does not suffer from these problems.  This problem may be addressed by fine-tuning SAMs decoder to produce class-level segmentation masks for certain classes where this is more appropriate.  

\section{Conclusions}\label{sec:conclusions}
In this work, we investigate whether the recently-proposed Segment Anything Model (SAM) \cite{kirillov2023segment} generalizes well to tasks involving overhead imagery.  We evaluated SAM on a large and diverse set of eight overhead imagery benchmark datasets.  For each benchmark, we investigated SAM's performance on two important potential application scenarios, in a similar fashion to \cite{kirillov2023segment}: model composition (COMP), and interactive segmentation (INTER).  We summarize our conclusions, labeled by the application area. 
\begin{itemize}
\item \textbf{(INTER)} SAM nearly always outperforms RITM (a baseline state-of-the-art annotation approach), and often by a large margin.  
\item \textbf{(INTER)} SAM achieves somewhat lower IOUs on overhead imagery compared to comparable experimental settings with natural imagery reported in \cite{kirillov2023segment}.  
\item \textbf{(INTER)} However, SAM can often achieve IoUs that would likely be satisfying for many practical annotation scenarios (e.g., SAM achieves >0.75 IoU on 6 of 10 benchmarks, with 10 point-based prompts). 
\item \textbf{(COMP)} SAM can often achieve performance that is comparable to out-of-the-box, but state-of-the-art, supervised models that have been trained on each task. However, its behavior is highly variable: in most cases it performs somewhat worse, but in some cases it performs superior to supervised models, while in others it performs much worse. \item \textbf{(COMP)} The aforementioned (COMP) finding pertain to bounding-box prompts provided by other vision models.  SAM generally performs poorly when prompted with point prompts obtained from other vision models. 
\item We also find that SAM is sensitive to the resolution of the imagery; e.g., artificially up-sampling the imagery can significantly improve or lower SAM's performance. 
\item SAM is designed to produce instance segmentations, however, this may not be well-suited to some object classes, such as roads, where we observe very poor performance.  
\end{itemize}

\textbf{Recommendations.} The performance of SAM can vary substantially on overhead imagery tasks.  This is typical of models being applied to novel data domains (i.e., overhead imagery instead of natural imagery), but users should be aware of this variability.  We suspect that fine-tuning SAM's decoder alone (as opposed to its larger encoder) may yield substantial improvements in its performance and reliability on overhead imagery.  In particular, retraining the decoder to better recognize common target class features (e.g., buildings, land-use) may be greatly beneficial.  Also, many tasks in overhead imagery are better-suited for class-level segmentation (e.g., roadways), and the decoder could be re-trained for class-level segmentation to substantially improve performance on road segmentation, or similar classes.  

{\small
\bibliographystyle{ieee_fullname}
\bibliography{egbib}
}


\end{document}